\documentclass[conference]{IEEEtran}
\IEEEoverridecommandlockouts
\usepackage{cite}

\usepackage{amsmath,amssymb,amsfonts}
\usepackage{algorithmic}
\usepackage{graphicx}
\usepackage{textcomp}
\usepackage{xcolor}
\usepackage[font=normal]{subcaption}
\usepackage[font=normal]{caption}
\def\BibTeX{{\rm B\kern-.05em{\sc i\kern-.025em b}\kern-.08em
    T\kern-.1667em\lower.7ex\hbox{E}\kern-.125emX}}
\begin{document}
\graphicspath{{./figures/}}
\title{Adaptive Ensembles of Fine-Tuned Transformers for LLM-Generated Text Detection
}

\author{\IEEEauthorblockN{Zhixin Lai}
\IEEEauthorblockA{\textit{Electrical and Computer Engineering } \\
\textit{Cornell University}\\
Ithaca, NY, USA \\
zl768@cornell.edu}
\and
\IEEEauthorblockN{Xuesheng Zhang}
\IEEEauthorblockA{\textit{Recommendation} \\
\textit{Meituan}\\
Beijing, China \\
xueshengz503@gmail.com}
\and
\IEEEauthorblockN{Suiyao Chen}
\IEEEauthorblockA{\textit{Industrial and Management Systems Engineering} \\
\textit{University of South Florida}\\
Tampa, FL, USA \\
suiyaochen@usf.edu}
}

\maketitle

\begin{abstract}
Large language models (LLMs) have reached human-like proficiency in generating diverse textual content, underscoring the necessity for effective fake text detection to avoid potential risks such as fake news in social media. Previous research has mostly tested single models on in-distribution datasets, limiting our understanding of how these models perform on different types of data for LLM-generated text detection task. We researched this by testing five specialized transformer-based models on both in-distribution and out-of-distribution datasets to better assess their performance and generalizability. Our results revealed that single transformer-based classifiers achieved decent performance on in-distribution dataset but limited generalization ability on out-of-distribution dataset. To improve it, we combined the individual classifiers models using adaptive ensemble algorithms, which improved the average accuracy significantly from 91.8\% to 99.2\% on an in-distribution test set and from 62.9\% to 72.5\% on an out-of-distribution test set. The results indicate the effectiveness, good generalization ability, and great potential of adaptive ensemble algorithms in LLM-generated text detection. 
\end{abstract}

\begin{IEEEkeywords}
LLM-generated text detection, Adaptive assemble algorithm, Transformer-based classifier, Generalization ability, in-distribution dataset, out-of-distribution dataset
\end{IEEEkeywords}

\section{Introduction}
Recently, LLM has experienced rapid development, and its text generation ability is comparable to that of human writing \cite{achiam2023gpt,anthropic2023modelcard}. LLM has penetrated into various aspects of daily life and plays a crucial role in many professional workflows such as forecasting and anomaly detection \cite{su2024large}, text production \cite{veselovsky2023artificial} and across various domains \cite{liu2024temperature,li2022tip,zhou2024visual,zhou2023thread}, promoting tasks such as advertising slogan creation, news writing \cite{liu2024news}, story generation, and code generation. In addition, their impact has significantly influenced the development of many sectors and disciplines, including education \cite{susnjak2022chatgpt}, law \cite{cui2023chatlaw}, biology \cite{piccolo2023many}, and medicine \cite{thirunavukarasu2023large}. However, the use of GPTs also brings various risks.

Firstly, GPT and other high-level language generation models can generate realistic text, which may be used to create and disseminate misleading information \cite{ji2023survey}, fake news, or harmful content. This may not only affect the public's understanding of the facts, but also manipulate key areas such as political elections, the stock market, and public health. Secondly, LLMs can generate text with a similar style to existing content, which may lead to copyright infringement and intellectual property disputes \cite{lee2023language}. For example, a model may replicate the style of a specific author in creating literary works, which may infringe upon the copyright of the original author. In addition, the content generated by LLMs may be used to produce pirated books, articles, or other media content, thereby damaging the economic interests and intellectual property rights of original content creators. Thirdly, LLM-generated content on social media platforms can be exploited to create false identities, enabling the manipulation of online conversations and public opinion. Automated accounts (bots) can leverage GPT-like models to generate a large volume of realistic comments, posts, or messages with deceptive, harassing, or persuasive intentions \cite{weidinger2021ethical,ayoobi2023looming}. In addition, the widespread use of models such as GPTs may lead to dishonest behavior in academic and educational fields. Students may use these tools to automatically generate papers and assignments, thereby undermining academic integrity \cite{stokel2022ai,kasneci2023chatgpt}. Lastly, many traditional industries are still cautiously adopting the power of LLM for text generation due to critical risk concerns such as cybersecurity \cite{liang2023study, tian2024lesson}, healthcare \cite{liu2023chatgpt, chen2019claims, li2024research, chen2017personalized, dai2023addressing, liu2023parameter}, transportation \cite{liu2024particle, chen2018data, tian2021adversarial} and sophistication and reliability requirements for manufacturing \cite{chen2017multi,wu2023genco,chen2020optimal,bingjie2023optimal}, environment \cite{chen2023sea}, agriculture \cite{tao2022optimizing,wu2022optimizing,wu2023extended} and energy \cite{shi2017combining}, etc.

However, telling machine-generated text from manually written text has proven to be a challenging task, with human performance only slightly surpassing chance levels \cite{mitchell2023detectgpt}. Consequently, the development of efficient automated methods for identifying machine-generated text and mitigating its potential misuse is important. Even prior to the widespread adoption of ChatGPT, research on the detection of machine-generated text had garnered attention, particularly in the recognition of deeply forged texts \cite{pu2023deepfake}. Nowadays, with ChatGPT's increasing prevalence, the primary focus of machine-generated text detection has shifted towards distinguishing between text generated by LLMs and text composed by humans \cite{jawahar2020automatic}. For instance, Guo et al. have undertaken the task of detecting whether a given text, in both English and Chinese, originates from ChatGPT or has been written by a human across various domains \cite{guo2023close}. 
While the use of pre-trained language models (LMs) as classifiers has been established as an effective approach for detecting text generated by Large Language Models (LLMs), prior research, often confined to single-model evaluations on familiar datasets, provides limited insights into their broader applicability and generalization capabilities. Individual classifiers may exhibit instability and struggle to generalize when applied to unfamiliar data contexts \cite{li2023deepfake}. In contrast, ensemble learning methods, such as the random forest algorithm \cite{breiman2001random}, excel in such situations by combining multiple models to create a more accurate and robust predictor. Recognizing this, we integrated ensemble learning algorithms to amalgamate our trained transformer-based classifiers, aiming to enhance their performance and generalization capacity.

The principal contributions of this work are as follows:
\begin{itemize}

    \item[$\bullet$]We achieved the LLMs generated text detection task by training five distinct transformer-based classifiers, each pre-trained on different datasets. . 

    \item[$\bullet$]We noticed variations in the accuracy of transformer-based classifiers and identified a notable constraint in their generalizability by evaluating the classifiers on different datasets.

    \item[$\bullet$]We employed both non-adaptive and adaptive ensemble techniques to improve the accuracy of LLM generated text detection. The adaptive ensemble method demonstrated the highest accuracy when tested on both in-distribution and out-of-distribution datasets, underscoring its superior effectiveness in identifying LLM-generated text. 
\end{itemize}

\section{Related Work}
The LLM-generated text detection task is framed as a binary classification problem. Transformer-based classifiers, which has been extensively explored in various domains \cite{dosovitskiy2020image, devlin2018bert, wu2024switchtab, chen2023recontab, li2024unlabeled}, serve as one of the most popular and effective approaches for this classification task. Additionally, we recognized that ensemble learning algorithms offer an efficient means of enhancing classification performance. In this paper, we combined individual classifiers with ensemble algorithms, resulting in a significant improvement in the performance of LLM-generated text detection on both in-distribution and out-of-distribution datasets.

\subsection{Algorithms for machine-generated text detection}

Guo et al, applied two methods: logistic regression with GLTR features and an end-to-end RoBerta classifier to distinguish whether a text was generated by ChatGPT or humans across multiple fields \cite{guo2023close}. Shijaku and Canhasi detected TOEFL essays using XGBoost with manually extracted 244 lexical and semantic features \cite{shijaku2023chatgpt}.
There are also widely-used off-the-shelf GPT detectors, such as the OpenAI detection classifier, GPTZero and ZeroGPT \cite{wang2023m4}. OpenAI's AI text classifier is fine-tuned on the output of an already trained language model. They used text generated by 34 models pre-trained by five different organizations, and then trained their models on samples from multiple sources of human writing and language model-generated text. GPTZero is trained on an extensive and diverse corpus of text created by humans and artificial intelligence, with a primary focus on English. As a classification model, GPTzero predicts whether a given text fragment is generated by a LLM with different text granularities, including sentence, paragraph, and entire document levels. These classifiers are all based on the transformer structure \cite{vaswani2017attention}, providing us with some reference for the algorithm of selecting classifiers.

\subsection{Ensemble Learning}
Ensemble learning refers to the machine learning paradigm where multiple learners (also known as models or predictors) are trained to solve the same problem. The main advantage of ensemble learning lies in its ability to improve model generalization ability. By combining multiple models, it can effectively reduce the risk of overfitting and underfitting. In addition, the diversity of different models can improve the overall prediction accuracy. Ensemble learning typically performs well in various machine learning competitions and practical applications \cite{zhu2024ensemble,hu2023leveraging}. 
We call algorithms without parameter updates in ensemble learning  as "non-adaptive ensemble algorithm", like the hard voting ensemble \cite{delgado2022semi}. And we call algorithms with parameter updates in ensemble learning as "adaptive ensemble algorithm", like the neural network ensemble and random forest algorithm. Usually, adaptive classifier detection performs better by adaptively integrating the outputs of different classifiers, assigning dynamic weights to each classifier's performance. 

\subsection{Ensemble Learning for LLM-generated Text Detection Task}
Specifically for LLM-generated text detection, since this is a unified data framework, ensemble learning will be a good fit to combine multiple models, instead of using fusion mechanisms \cite{weimin2024enhancing,li2023joyful, wang2023emp} or model selection techniques \cite{hu2023many}.
LLM-Blender \cite{friedman2001greedy} is an ensemble framework designed to attain consistently superior performance by leveraging the diverse strengths of multiple open-source large language models (LLMs). The study \cite{abburi2023generative} presents ensemble neural models utilizing probabilities from multiple pre-trained Large Language Models (LLMs) as features for Traditional Machine Learning (TML) classifiers to distinguish between AI-generated and human-written text, achieving competitive performance in both English and Spanish languages and ranking first in model attribution.

\section{Dataset}
We utilized the DAIGT dataset for training and in-distribution testing, consisting of a 2:1 ratio of human-generated to LLM-generated text. The LLM-generated text includes outputs from various LLMs, such as ChatGPT and Llama-70b \cite{kleczek2013daigt}. We divided the dataset into training and testing sets in an 80\%/20\% ratio, as shown in Table~\ref{datasets_item_number}.

\begin{table}[htbp] \renewcommand{\arraystretch}{1.5}\addtolength{\tabcolsep}{-4.6pt}
\caption{The item number of training and testing datasets}
\footnotesize
\begin{center}
\begin{tabular}{|c|c|c|c|c|}
\hline
\textbf{Dataset}&\textbf{Functionality}&\textbf{Human-generated}&\textbf{LLM-generated}&\textbf{Total}\\
\hline
{DAIGT}&{Training}&{23833}&{11531}&{35364}\\
\hline
{DAIGT}&{In-dist.$^{\mathrm{a}}$ testing}&{5959}&{2883}&{8842}\\
\hline
{Deepfake}&{Out-of-dist.$^{\mathrm{a}}$ testing}&{800}&{762}&{1562}\\
\hline 
\multicolumn{5}{l}{$^{\mathrm{a}}$dist.: distribution}
\end{tabular}
\label{datasets_item_number}
\end{center}
\end{table}

\begin{figure}[h!]
	\centering
	\begin{subfigure}[b]{0.80\linewidth}
		\centering
		\includegraphics[width=\linewidth]{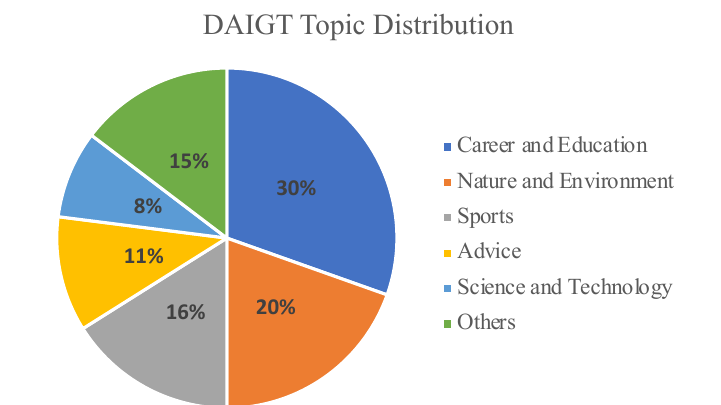}
		\label{DAIGT}
	\end{subfigure}
	\begin{subfigure}[b]{0.80\linewidth}
		\centering
		\includegraphics[width=\linewidth]{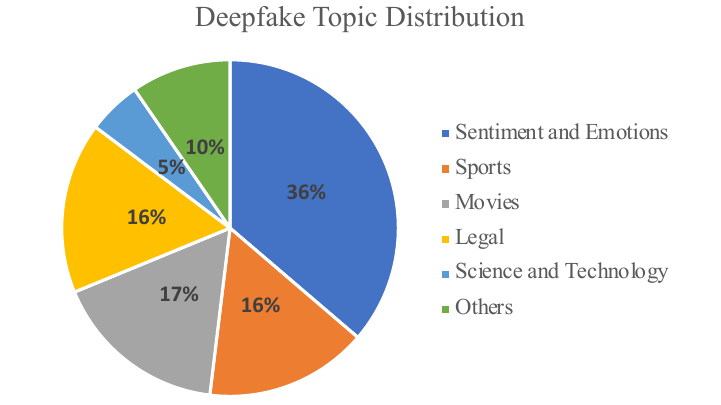}
		\label{Deepfake}
	\end{subfigure}
	\vspace{-2.2ex}
	\caption{Datasets topic distribution}  
	\label{Topic_Distribution}
\end{figure}

\begin{figure*}[htbp]
\centerline{\includegraphics[width=1\linewidth]{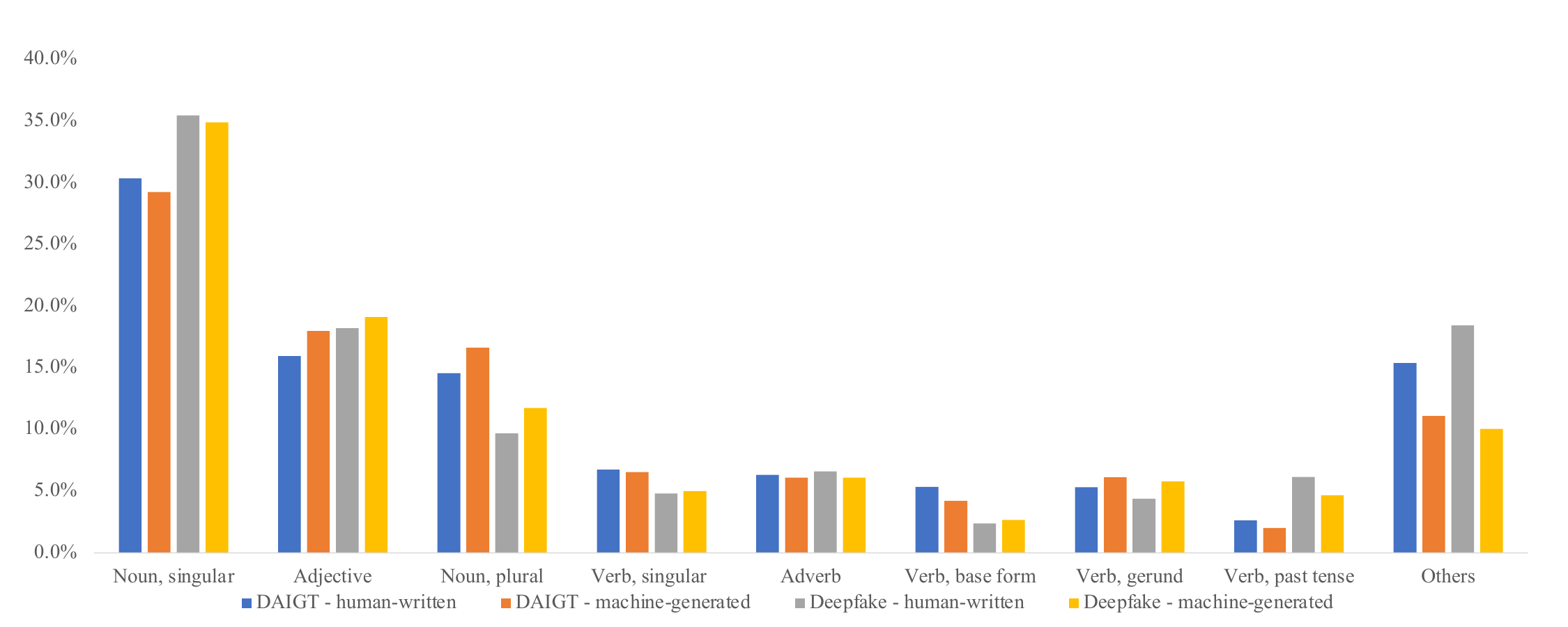}}
\caption{Dataset part-of-speech tag.}
\label{part_of_speech}
\end{figure*}

\begin{table}[htbp] \renewcommand{\arraystretch}{1.5}\addtolength{\tabcolsep}{10pt}
\caption{Test average word length of training and testing Datasets}
\footnotesize
\begin{center}
\begin{tabular}{|c|c|c|}
\hline
\textbf{Dataset}&\textbf{Human-written}&\textbf{Machine-generated}\\
\hline
{DAIGT}&{466.82}&{369.09}\\
\hline
{Deepfake}&{279.43}&{284.33}\\
\hline
\end{tabular}
\label{datasets_word_length}
\end{center}
\end{table}
To assess the generalizability of our text detection methods, we introduced the Deepfake dataset as out-of-distribution test set. The Deepfake, generated by a range of LLMs, encompasses broader domains including open statements, news articles, and scientific texts \cite{li2023deepfake}. 
We compared the two datasets from the following aspects: 
\begin{itemize}
    \item[$\bullet$]Firstly, the DAIGT dataset has longer text lengths than the Deepfake dataset from Table~\ref{datasets_word_length}. 

    \item[$\bullet$]Secondly, we used the Latent Dirichlet Allocation (LDA) to analyze the dataset topic distribution, shown in Fig.~\ref{Topic_Distribution}. The topic distribution indicates the dissimilarity in topic type and distribution between DAIGT and Deepfake datasets, indicating their divergence in content. 

    \item[$\bullet$]Thirdly, from Fig.~\ref{part_of_speech}, part-of-speech distribution in human-written and machine-generated text within the same dataset closely aligns, suggesting similar linguistic structures. However, when we compare DAIGT with Deepfake, a more notable disparity emerges in their part-of-speech distribution. 
\end{itemize}

The analyses from three different aspects of text length, topic, and part-of-speech distribution shows great difference between the DAIGT and Deepfake datasets. Therefore, we are confident to use Deepfake as an out-of-distribution dataset to test generalizability of different models.

\begin{figure}[htbp]
\centerline{\includegraphics[width=1\linewidth]{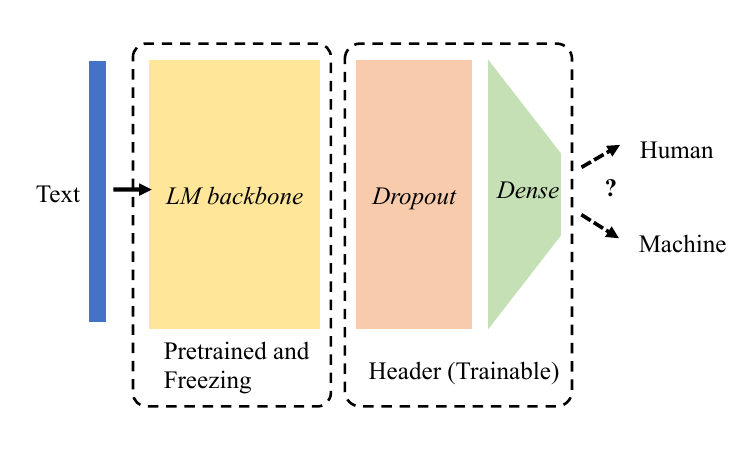}}
\caption{Structure of single classifer detection.}
\label{single_classifier_detection}
\end{figure}

\begin{figure*}[htbp]
\centerline{\includegraphics[width=1\linewidth]{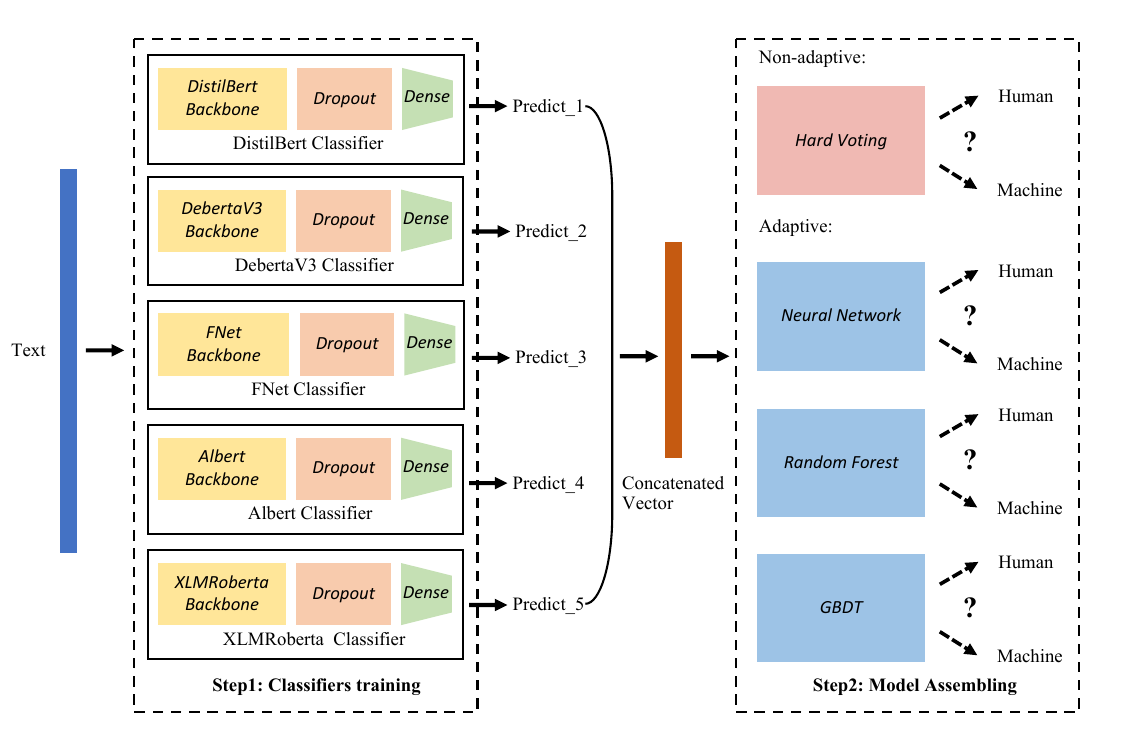}}
\caption{Structure of assemble detection.}
\label{assemble_detection}
\end{figure*}

\begin{figure}[htbp]
\centerline{\includegraphics[width=1\linewidth]{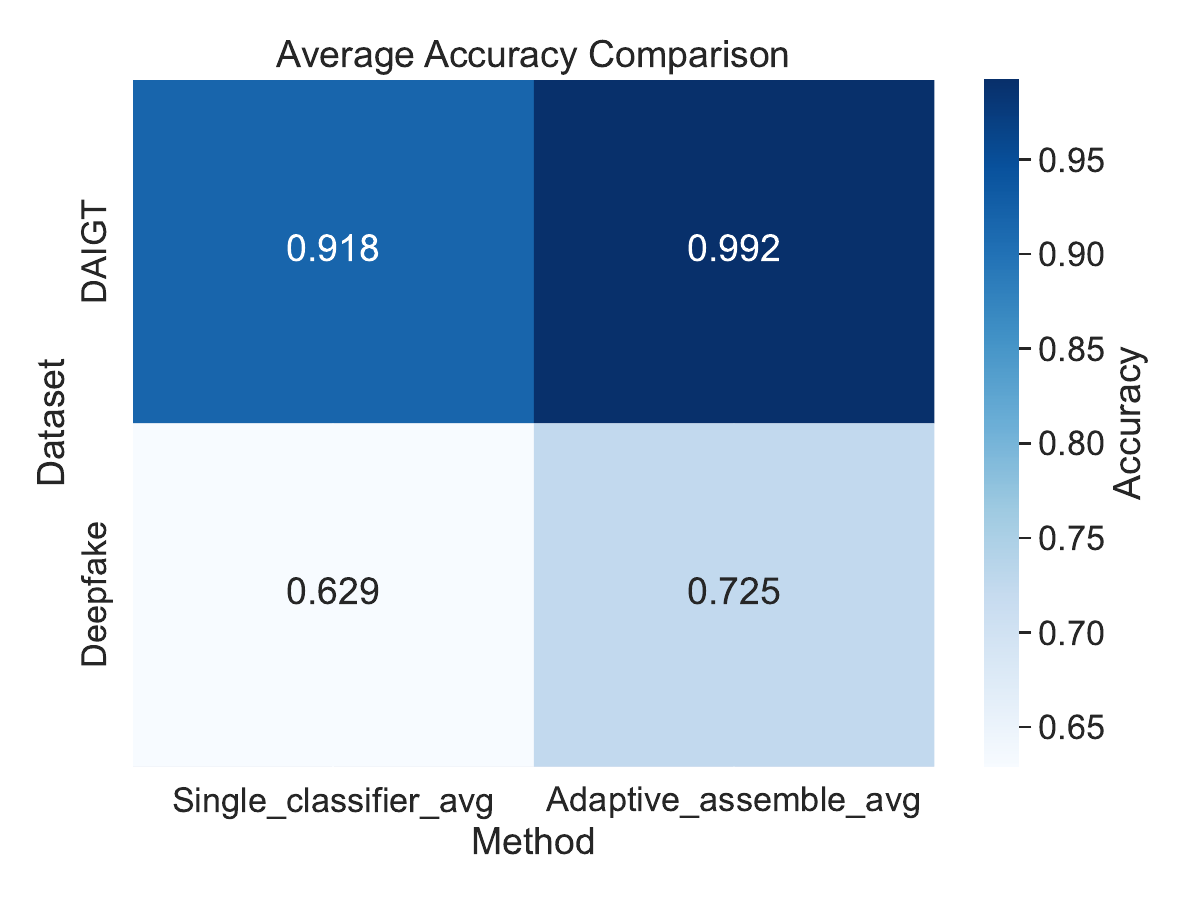}}
\caption{Average accuracy of different methods.}
\label{average_accuracy}
\end{figure}

\begin{table*}[htbp]\renewcommand{\arraystretch}{1.5}\addtolength{\tabcolsep}{5pt}
\caption{Detailed detection performance on in-distribution dataset}
\begin{center}
\begin{tabular}{|c|c|c|c|c|c|c|c|c|}
\hline
&&\multicolumn{7}{|c|}{\textbf{Metrics}}\\
\hline
Strategy&\textbf{Method}&\multicolumn{3}{|c|}{\textbf{Human-generated text}}&\multicolumn{3}{|c|}{\textbf{LLM-generated text}}&\textbf{Global}\\
\hline
&&\textbf{Recall}&\textbf{Precision}&\textbf{F1}&\textbf{Recall}&\textbf{Precision}&\textbf{F1}&\textbf{Accuracy}\\
\cline{2-9}
&DistilBert &{0.863}&\textbf{0.997}&{0.925}&\textbf{0.994}&{0.780}&{0.875}&{0.906}\\
\cline{2-9}
&DeBERTaV3 &{0.967}&{0.995}&{0.981}&{0.990}&{0.936}&{0.962}&{0.974}\\
\cline{2-9}
Single Classifier&FNet &{0.961}&{0.862}&{0.909}&{0.685}&{0.896}&{0.776}&{0.870}\\
\cline{2-9}
&Albert &{0.959}&{0.929}&{0.943}&{0.85}&{0.909}&{0.879}&{0.923}\\
\cline{2-9}
&XLMRoberta &{0.879}&{0.996}&{0.934}&{0.993}&{0.801}&{0.887}&{0.917}\\
\hline
Non-adaptive Ensemble&Hard Voting &{0.960}&{0.995}&{0.977}&{0.991}&{0.924}&{0.956}&{0.970}\\
\hline
&Neural Network &{0.996}&{0.992}&\textbf{0.994}&{0.984}&{0.992}&\textbf{0.988}&\textbf{0.992}\\
\cline{2-9}
Adaptive Ensemble&Random Forest &\textbf{0.997}&{0.992}&\textbf{0.994}&{0.983}&\textbf{0.993}&\textbf{0.988}&\textbf{0.992}\\
\cline{2-9}
&GBDT &{0.995}&{0.992}&{0.993}&{0.983}&{0.990}&{0.987}&\textbf{0.992}\\
\hline
\end{tabular}
\label{detection_performance}
\end{center}
\end{table*}

\begin{table}[htbp]\renewcommand{\arraystretch}{1.5}\addtolength{\tabcolsep}{-4pt}
\caption{Accuracy Comparison between in-distribution and out-of-distribution datasets}
\begin{center}
\begin{tabular}{|c|c|c|c|}
\hline
&&\multicolumn{2}{|c|}{\textbf{Accuracy}}\\
\cline{3-4}
&Method&In-dist. testing& Out-of-dist. testing\\
&&DAIGT& Deepfake\\
\hline
&DistilBert &{0.906}&{0.567}\\
\cline{2-4}
&DeBERTaV3 &{0.974}&{0.616}\\
\cline{2-4}
Single Classifier&FNet &{0.870}&{0.659}\\
\cline{2-4}
&Albert &{0.923}&{0.699}\\
\cline{2-4}
&XLMRoberta &{0.917}&{0.602}\\
\hline
Non-adaptive Ensemble&Hard Voting &{0.970}&{0.651}\\
\hline
&Neural Network &\textbf{0.992}&\textbf{0.736}\\
\cline{2-4}
Adaptive Ensemble&Random Forest &\textbf{0.992}&\textbf{0.722}\\
\cline{2-4}
&GBDT &\textbf{0.992}&{0.718}\\
\hline
\end{tabular}
\label{Accuracy_Comparison}
\end{center}
\end{table}

\section{Method}
In this study, we firstly developed single classifier models leveraging pretrained transformers. Also, we aggregated the output of the classifiers with non-adaptive ensemble and adaptive ensemble methods. The performance and generalizability of these models and methods was subsequently evaluated on DAIGT and Deepfake datasets.
\subsection{Single Classifier Detection}
The classifiers are pre-trained transformer-based LMs, which were subsequently fine-tuned for the LLM-generated text detection task. As illustrated in Fig.~\ref{single_classifier_detection}, each classifier is associated with a classification head, comprising dropout and dense layers, integrated with the backbone of a transformer-based LM. We employed five distinct transformer-based LMs as the backbones for five individual classifiers. The pre-trained layers of the LM are frozen (not updated during training) to preserve their learned features. Only the classifier “head” (the last layer) is trained. 
In addition, we use cross-entropy loss and Adam optimizer with a learning rate of 5e-4. During model training, each text sample is truncated to a maximum of 256 tokens, the classifier is trained for 8841 steps and each training batch contains 4 samples.

The summary of the pre-trained LMs is provided below:
1)	DistilBert: DistilBERT is developed through the distillation of the BERT base model, is notably smaller and faster \cite{sanh2019distilbert}. We use the ``distil\_bert\_base\_en\_uncased" as pretrained backbone weight. 
2)	DeBERTaV3: DeBERTaV3 is an enhanced pre-trained language model that outperforms its predecessor by implementing replaced token detection (RTD) and introducing a gradient-disentangled embedding sharing method \cite{he2021debertav3}. We use the ``deberta\_v3\_base\_en" as pretrained backbone weight.
3)	FNet: FNet introduces a novel approach to speed up Transformer encoders by replacing self-attention sublayers with linear transformations, achieving 92-97\% of BERT's accuracy on the GLUE benchmark while training significantly faster \cite{lee2021fnet}. We use the ``f\_net\_base\_en" as pretrained backbone weight.
4)	Albert: Albert presents two parameter-reduction techniques that lower memory usage and accelerate training of BERT, achieving superior scaling and performance \cite{lan2019albert}. We use the ``albert\_base\_en\_uncased" as pretrained backbone weight.
5)	XLMRoberta: XLMRoberta is a large-scale multilingual transformer-based LM pretrained on a diverse set of one hundred languages using extensive CommonCrawl data \cite{conneau2019unsupervised}. We use the ``xlm\_roberta\_base\_multi" as pretrained backbone weight.

\subsection{Non-Adaptive Ensemble}

1)	Hard Voting Ensemble: The hard voting ensemble integrates the outputs of the distinct classifiers through a hard voting mechanism. Shown in Fig.~\ref{assemble_detection}, this ensemble technique operates by aggregating the predictions from each individual classifier without extra model training, where each classifier contributes a “vote” towards the final decision. The prediction receiving the majority of votes is then selected as the ensemble's output.
\subsection{Adaptive Ensemble}
Adaptive ensemble presents adaptive approaches of classifier aggregation, including neural network ensemble random forest ensemble, and GBDT ensemble. Shown in Fig.~\ref{assemble_detection}, the adaptive ensemble methods assign dynamic weights to the output of each classifier based on their performance, enabling a more effective and context-sensitive combination of predictions. 
1)	Neural Network Ensemble: The neural network ensemble integrates with two ReLU-activated dense layers of 32 and 16 neurons, each followed by 50\% dropout for regularization, following the output from the five transformer-based classifiers, shown in Fig.~\ref{assemble_detection}. This structure employs dense layers to effectively synthesize the classifiers' outputs, while the dropout layer aids in preventing overfitting, thus enhancing the model's generalization capabilities. During the model training for the Neural Network, the five classifiers are frozen. The model was trained for 10 epochs using a batch size of 128 with the Adam optimizer and cross-entropy loss. The training process took just around 5 seconds.
2)	Random Forest Ensemble: The random forest algorithm \cite{breiman2001random} with 100 estimators is used to adaptively aggregate the outputs of the distinct classifiers, shown in Fig.~\ref{assemble_detection}. During the model training for Random Forest, the five classifiers are frozen. This method capitalizes on the inherent adaptability and decision-making prowess of random forests to dynamically integrate classifier outputs, thereby enabling a more nuanced and context-sensitive ensemble decision-making process. 
3)	GBDT (Gradient Boosting Decision Trees) Ensemble: GBDT [31] with 100 estimators adaptively aggregates outputs from the five classifiers, shown in Fig.~\ref{assemble_detection}. During the model training for the GBDT, the five classifiers are frozen. The approach ensures a dynamic and context-sensitive integration of diverse classifier predictions.

\section{Results}
We employed the standard metrics conventionally utilized in the domain of text classification: recall, precision, F1, and accuracy \cite{sebastiani2002machine}. 
From Table~\ref{detection_performance} and \ref{Accuracy_Comparison}, the adaptive ensemble methods have superior performance compared to single classifier models and non-adaptive ensemble. The adaptive ensemble methods achieve the best F1 scores for both LLM-generated and human-written texts, as well as the top accuracy across both in-distribution and out-of-distribution datasets. The findings highlight the adaptive ensemble methods' robustness and their enhanced ability to generalize in detecting LLM-generated text.

\subsection{Single Model Detection}
As indicated in Table~\ref{detection_performance}, the single classifiers demonstrate a predilection for higher Precision and F1 scores when classifying human-generated text, mainly attributed to the imbalanced dataset. The recall rates for human-generated versus LLM-generated text differ significantly among classifiers, underscoring the distinctiveness of each model. This diversity suggests that these models are well-suited for combination through ensemble algorithms, potentially enhancing overall classification performance. 
In Table~\ref{Accuracy_Comparison}, compared to in-distribution test set DAIGT, there is a significant accuracy regression on out-of-distribution test set Deepfake. Single classifier has bad generalization capabilities for unseen dataset. Also, DeBERTaV3 performs best in DAIGT while Albert performs best in Deepfake, which approves the generalization capabilities variance among transformer-based classifiers. 

\subsection{Non-Adaptive Ensemble Detection}
The hard voting ensemble method achieved an accuracy of 0.970, surpassing most single classifier models, though it slightly underperforms the highest-performing DeBERTaV3 classifier, which recorded an accuracy of 0.974 on in-distribution dataset. This outcome aligns with expectations, as hard voting is a non-adaptive ensemble approach, and its performance is typically near that of the best-performing individual model. 
\subsection{Adaptive Ensemble Detection}
In Table~\ref{detection_performance}, the adaptive ensemble detection approach achieved an accuracy of 0.992 on the in-distribution test set, outperforming single classifier models and non-adaptive ensemble method. It also achieved the highest F1 scores for both detecting human and LLM-generated texts, demonstrating superior classification capabilities. 
Table~\ref{Accuracy_Comparison} shows that adaptive ensemble methods also excel on out-of-distribution datasets, with the Neural Network Ensemble reaching the highest accuracy of 0.736, evidencing better generalization capabilities than single transformer-based classifier. 
In Fig.~\ref{average_accuracy}, adaptive assemble methods have the average accuracy of 0.992 on DAIGT and 0.725 on Deepfake, while transformer-based classifiers have 0.918 on DAIGT and 0.629 on Deepfake. 
In summary, the adaptive ensemble methods, integrated with its constituent transformer-based classifiers, significantly enhances the performance and generalization ability of LLM-generated text detection task, based on the results cross different datasets. 

\section{Conclusion}
This paper has successfully demonstrated the significant advantages of adaptive ensemble methods in the field of text classification in distinguishing between text generated by humans and large language models. This paper revealed that while individual transformer-based classifiers can tell LLM generated text, they exhibit constraints in their capacity to generalize when confronted with out-of-distribution data. The implementation of adaptive ensemble strategies has proven to be instrumental in mitigating the limitation. The adaptive ensemble not only improved accuracy in in-distribution dataset significantly but also showed better generalization ability out-of-distribution dataset. This dual enhancement in accuracy and generalizability makes the adaptive ensemble method as a robust, excellent tool in the ongoing challenge of LLM-generated text detection.
\section*{Acknowledgment}
Sincere thanks to Zelun Wang for paper review.
\bibliographystyle{IEEEtran}
\bibliography{main}

\end{document}